\documentclass[conference]{IEEEtran}
\IEEEoverridecommandlockouts
\usepackage{cite}
\usepackage{subfigure}
\usepackage[algoruled, resetcount, linesnumbered]{algorithm2e}
\usepackage{amsmath,amssymb,amsfonts}
\usepackage{algorithmic}
\usepackage{graphicx}
\usepackage{textcomp}
\usepackage{xcolor}
\def\BibTeX{{\rm B\kern-.05em{\sc i\kern-.025em b}\kern-.08em
    T\kern-.1667em\lower.7ex\hbox{E}\kern-.125emX}}
\begin{document}

\title{Unsupervised Anomalous Trajectory Detection for Crowded Scenes}

\author{\IEEEauthorblockN{Deepan Das}
\IEEEauthorblockA{\textit{Dept. of Electronics and Telecommunication Engineering} \\
\textit{Indian Institute of Engineering Science and Technology, Shibpur}\\
WB,India \\
ddas27@wisc.edu}
\and
\IEEEauthorblockN{Deepak Mishra}
\IEEEauthorblockA{\textit{Dept. of Avionics} \\
\textit{Indian Institute of Space Science and Technology}\\
Thiruvananthapuram, India \\
deepak.mishra@iist.ac.in}
}

\maketitle

\begin{abstract}
We present an improved clustering based, unsupervised anomalous trajectory detection algorithm for crowded scenes. The proposed work is based on four major steps, namely, extraction of trajectories from crowded scene video, extraction of several features from these trajectories, independent mean-shift clustering and anomaly detection. First, the trajectories of all moving objects in a crowd are extracted using a multi feature video object tracker. These trajectories are then transformed into a set of feature spaces. Mean shift clustering is applied on these feature matrices to obtain distinct clusters, while a Shannon Entropy based anomaly detector identifies corresponding anomalies. In the final step, a voting mechanism identifies the trajectories that exhibit anomalous characteristics. The algorithm is tested on crowd scene videos from datasets. The videos represent various possible crowd scenes with different motion patterns and the method performs well to detect the expected anomalous trajectories from the scene.
\end{abstract}


\section{Introduction}
Computer Vision research aims to converge at human-like abilities to interpret and extract useful information regarding behavioural patterns and anomalies from a descriptive set of visual data. However, human abilities have glaring limitations when it comes to analyzing simultaneously changing signals\cite{sulman2008effective}. A crowd presents itself as a considerably large collection of simultaneously changing parameters, characterized by usual dominant patterns and some observable abnormalities. Safety is the primary reason to understand crowd dynamics and isolate anomalous patterns. With crowd-related violent incidents on the rise, it is paramount that we expand our studies to analyze the intricate and complex nature of crowds. Understanding anomalies in a crowded scene enables better public space design and also allows better surveillance systems to be built. Earlier works like those of Kim et al.\cite{kim2009observe} used a Mixture of Probabilistic Principal Component Analyzers to learn patterns of local optical flow and then validate the consistency by Markov Random Field. Cong et al.\cite{cong2013abnormal} used a multi-scale histogram of Optical Flow as the feature descriptor and used it as the basis for a sparse reconstruction. Ali et al.\cite{ali2007lagrangian} used Lagrangian Particle Dynamics to model coherent crowd flow as fluid flow.

\begin{figure}[htbp]
	\subfigure[Snapshot from the Pilgrim sequence of the UCF Database of crowded scenes]
	{\label{fig:edge-a}
		\includegraphics[scale=0.11]{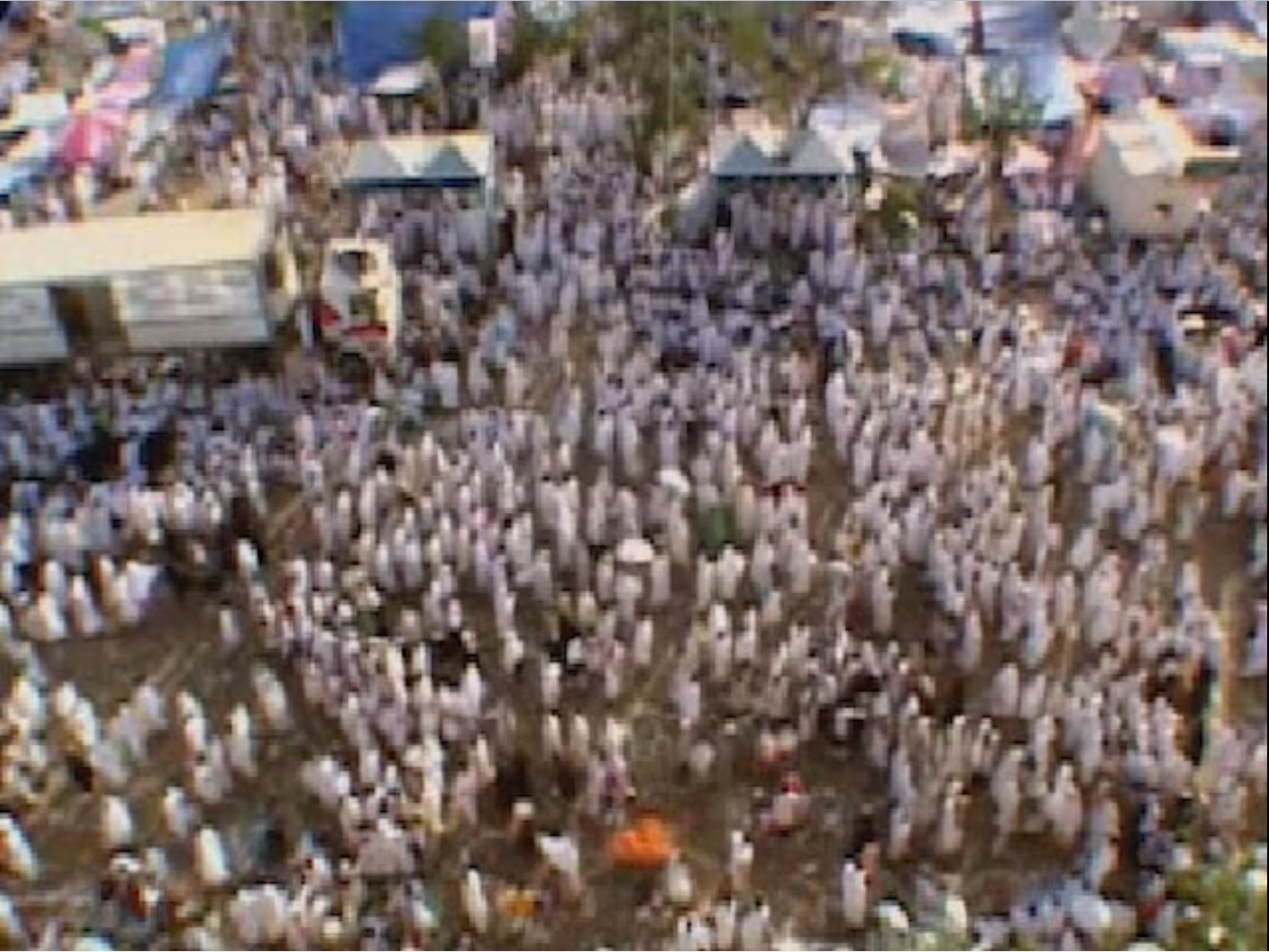}}
	\subfigure[Snapshot from the Crowded Intersection sequence of the UCF Database of crowded scenes]
	{\label{fig:edge-b}
		\includegraphics[scale=0.11]{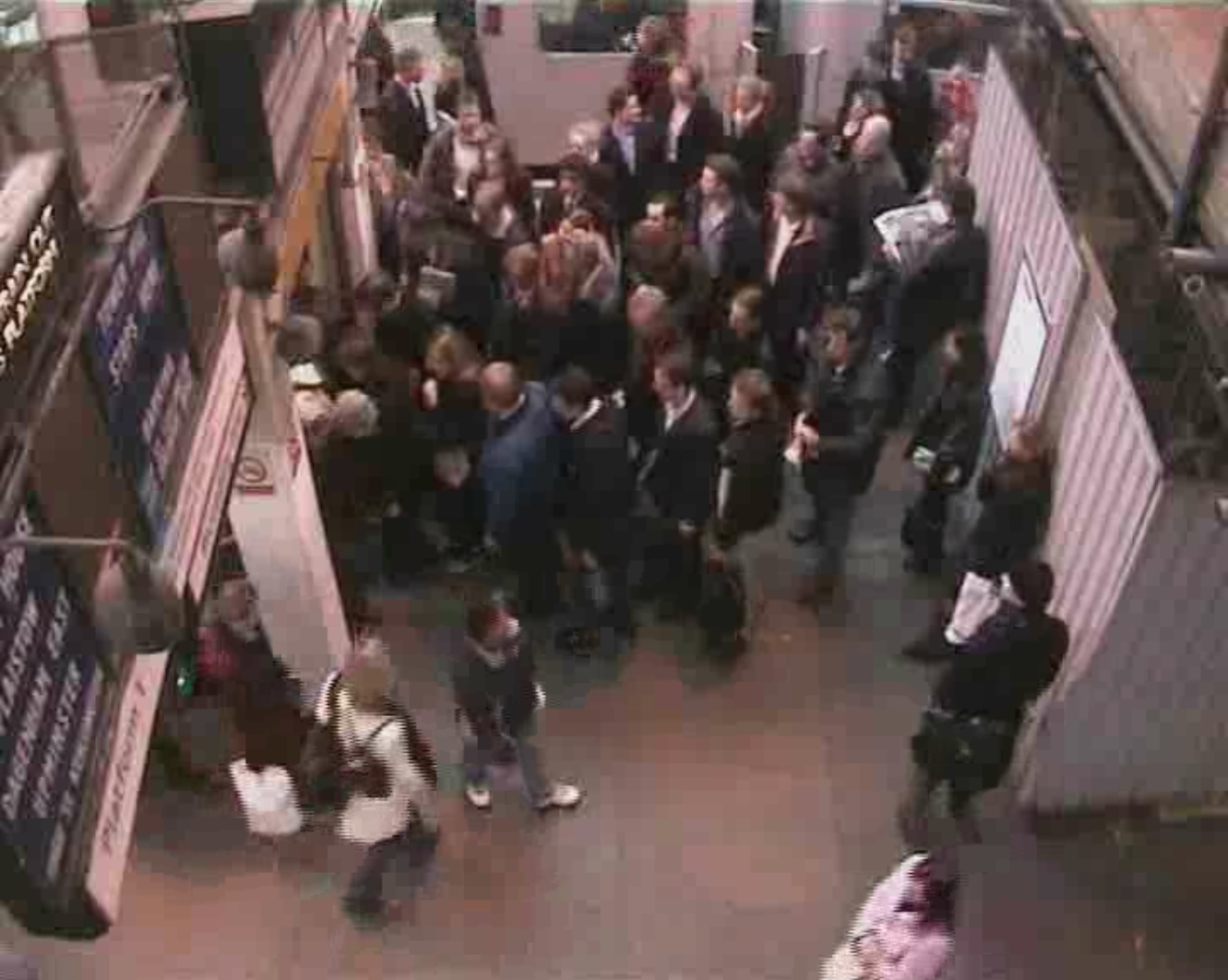}}
	\caption{Typical Crowded Scenes}
	\label{fig:edge}
\end{figure}

In general, Supervised methods require a considerable amount of labeled data, which is directly utilized to build the connection between video features and video labels. Therefore, developing Unsupervised anomaly detection systems prove to be more challenging than supervised ones. An anomaly in a crowded scene can be determined from the motion patterns of it's constituent pedestrians and objects. Analyzing trajectory data enables one to predict and identify anomalies with an excellent degree of accuracy. The early works on trajectory analysis includes that of Fu et al.\cite{fu2005similarity} which proposed a hierarchical  clustering framework to classify vehicle motion trajectories based on pairwise similarities, but with the limitation of using only a single feature for clustering. Progressing further, Anjum and Cavallaro\cite{anjum2008multifeature} proposed the use of multiple features in a Mean shift clustering based framework. They could identify outliers using a basic mean trajectory location based measure. Antonini et al.\cite{antonini2006counting} transformed the input trajectories using Independent Components Analysis and then use Euclidean distance to find similarities between various trajectories. The Shannon Entropy measure has presented itself as an excellent tool for many applications including video key selection\cite{xu2014browsing}, Network anomaly detection\cite{santiago2015entropy} and worm detection\cite{ranjan2007dowitcher}. The principal contributions of this paper include the incorporation of a multi-feature object tracker that works excellently well for crowded scenes\cite{Sharma2016ATC} and the use of multiple features for independent clustering. Furthermore, an information theory based Shannon entropy measure is proposed to detect anomalies for each cluster and then identify overall anomalous trajectories for the entire scene using a voting mechanism. The paper is organized as follows: Section II discusses the trajectory estimation and feature extraction procedure. Section III discusses the Clustering task with Section IV focusing on the Anomaly detection mechanism while Section V sheds light on the results obtained using the algorithm. 
%

\section{Trajectory and Feature Extraction}
The first task is to evaluate trajectory paths for all moving objects. 

\subsection{Trajectory Extraction}
The estimation of trajectories in crowded scenes is a challenging task due to various factors like high degree of occlusion, difficulty in tracking individual objects and arbitrary changes in nature of the motion. To tackle this problem, we incorporate the use of a multi-object tracker, that works exceedingly well in crowded scenes as demonstrated by Sharma et al.\cite{Sharma2016ATC}. Using this approach, each frame is divided into non-overlapping boxes and low-level features are detected inside each box. Following this, the centroids of all the detected feature points in each box is tracked using the standard Kanade Lucas tracking algorithm. Fresh boxes are introduced periodically to track newly introduced objects.

\subsection{Feature Extraction}
Most trajectory clustering and anomaly classifiers used a single feature descriptor for the task. We propose the use of multiple features, namely: 

\subsubsection{Density}
A trajectory can have varying densities around it, depending on the size of it's neighbourhood. The density feature is thus computed using varying sizes of neighbourhood $\epsilon$. We have considered three varying sizes as proposed by Sharma et al.\cite{Sharma2016ATC}. 
\begin{equation}
n_{T,j, \epsilon} = |\{ T^{i} | \forall i \neq j, d(f^{j},f^{i})<\epsilon\} |
\end{equation}
\begin{equation*}
F^{j} = [ n_{j,\epsilon1}, n_{j,\epsilon2}, n_{j,\epsilon3} ]
\end{equation*}
In this work, we are also interested in distances that describe the similarity of objects along time and therefore are computed by analysing the way distance between the objects varies over time. This gives us a measure of the \emph{spatio-temporal density} in the most natural way possible:
\begin{equation}
D(\tau_{1}, \tau_{2})|_{T} = \frac{\int d(\tau_{1}(t), \tau_{2}(t))dt }{|T| }
\end{equation}
Where $ d(\tau_{1}(t), \tau_{2}(t)$ represents the pairwise distance between two trajectories at the instant $t$. 

\subsubsection{Shape}
All trajectory sketch a particular shape across the spatio-temporal scene, and this is represented as a polynomial function. The coefficients are calculated separately for the $x$ and $y$ coordinates yielding the $f_s$ feature vector.
\begin{eqnarray}
x(t) = a_0 +a_1t+a_2t^2+a_3t^3\\
y(t) = b_0 +b_1t+b_2t^2+b_3t^3
\end{eqnarray}
\begin{equation*}
f_s = [a_0,\dots,a_3,b_0,\ldots,b_3]
\end{equation*}
\begin{figure}[htbp]
	\subfigure[Snapshot from the Crowded Subway exit sequence]
	{\label{fig:edge-a}
		\includegraphics[scale=0.12]{typicalcrowd1.PNG}}
	\subfigure[Extracted trajectories. Green dots are starting points]
	{\label{fig:edge-b}
		\includegraphics[width=4cm]{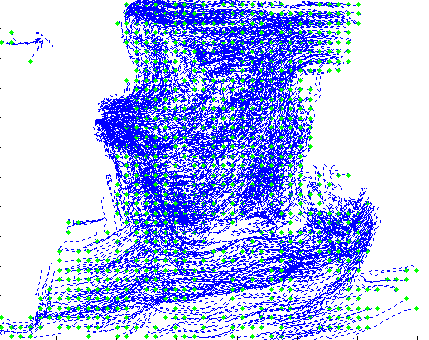}}
	\caption{Crowded scene and extracted trajectories}
	\label{fig:2}
\end{figure}
\subsubsection{Mean Position}
It may be possible that trajectories separated over large distances may have similar velocities, directions and density features and consequently, get clustered in the same group. To avoid this, a location measure is needed as $\mathnormal{f_l = [mean_x, mean_y] } $. 
\subsubsection{Standard Deviation}
Standard Deviation is an extremely popular measure that quantifies the amount of variation or dispersion in a time-series data. 
\begin{equation}
\sigma = \sqrt{(E[(X-\mu)^2])}
\end{equation}
The trajectories extracted from each surveillance video will give rise to a distinct feature-space for each of the features mentioned above. These distinct feature spaces will be used for identifying anomalies for that particular feature, and thereafter, the detection of overall anomalies.

\section{Clustering}
Clustering methods have gained immense popularity as a data analysis tool ever since Clements\cite{clements1954use} introduced it in 1954. It is observed that significantly dominant and usual features correspond to the denser regions of the probability density function of the data points. Using a Kernel Density Estimate, the modes of the probability density function can be found using either the Mean Shift\cite{fukunaga1975estimation,cheng1995mean} or the Mountain method\cite{yager1994approximate}. We would be using the Mean Shift method here as proposed by Fukunaga and Hostetler\cite{fukunaga1975estimation}. Moreover, since the anomaly detection algorithm proposed here revolves around clustering similar data points, the clustering algorithm used here has to be highly effective, as demonstrated by the Mean Shift Clustering algorithm.

\subsection{Mean Shift Clustering}
It is a non-parametric and versatile, iterative algorithm with applications in varied fields like object tracking, texture segmentation and data mining. After learning estimate of the probability density of the data points using a Kernel Density Estimate, a gradient ascent procedure associates each data point with the nearby peak of the data-set's density function. It defines a window around it and computes the mean of all the data-points within the window and shifts the centre of the window to the new mean until the process converges. When the process converges, we obtain the modes of the density estimate which serve as the centre-points of the clusters in the data. Suppose, there are $n$ data-points in the d-dimensional space $\mathcal{R}_d$, then the density estimate with Kernel $K(x)$ and bandwidth $h$, can be denoted as $ f_{h,k}(x)$. If we define $g(x) = -\acute{K}(x)$ as a shadow function\cite{wu2007mean} of $K(x)$, with the assumption that the derivative of the kernel $K$ exists for all $x \in [0,\infty)$, then the gradient of the density estimate can be written as:
\begin{equation}
\nabla f_{h,k}(x) = \frac{2c_{k,d}}{nh^{d+2}} \sum_{i=1}^{n}  (x_{i} - x)\acute{K}( || \dfrac{x-x_{i}}{h} ||^{2})
\end{equation}
\begin{equation}
\nabla f_{h,k}(x) =\\
\dfrac{2c_{k,d}}{nh^{d+2}}[ \sum_{i=1}^{n} g(|| \dfrac{x-x_i}{h} ||^2) ]\textbf{m}_{h,G}(x) 
\end{equation}
The modes of the density function are obtained among the zeros of the gradient of the density function. The first term in the product is proportional to the density estimate at $x$ computed with kernel $G$, while the second term, or the \emph{mean shift} is defined as the difference between the weighted mean and the centre of the Kernel window. 
\begin{equation}
\textbf{m}_{h,G}(x) = [ \dfrac{\sum_{i=1}^{n} x_{i}g(|| \frac{x-x_{i}}{h} ||^{2})}{\sum_{i=1}^{n} g( || \frac{x-x_{i}}{h} ||^{2}) } - x]
\end{equation}
It can be observed that the mean shift vector always points towards the direction of maximum increase in the density\cite{comaniciu2002mean}. These obtained modes, or cluster centres, are found for each independent feature obtained, therefore, giving us a non-overlapping set of trajectories that are characteristic of the cluster they belong to.

\begin{algorithm}
	\SetAlgoLined
	\KwIn{Crowded Video Sequence}
	\KwOut{Anomalous Trajectories in the Sequence}
	Extract Trajectories from Video\\
	Compute Features $F_1, F_2,\ldots,F_n$\\
	\For{$i = 1$ to $n$}{
		Compute Cluster Centres for $F_i$\\
		$C_i = [c_{i,1}, c_{i,2},\dots, c_{i,k}]$\\
		\For{$i = 1$ to $numTrajec$}{
			\For{$j=1$ to $k$}{
				$distvec(l,j) = dist(trajec_l, c_{i,j})$\\
			}
		}
		\For{$i = 1$ to $numTrajec$}{
			\For{$j=1$ to $k$}{
				$P(i,j) = distvec(i,j)/ \sum\limits_{m=1}^{k}distvec(i,m)$\\
			}
		}
		\For{$i = 1$ to $numTrajec$ }{
			$H(i) = - \sum_{j=1}^{k} P_{i,j} log P_{i,j}$\\
			\uIf{$H(i) > thresh$}{
				Vote(i)++\\
			}
		}
	}
	\uIf{$Vote(i) > n/2$}{
		$Anom(i) = 1$
	}
	\Return $Anom$ 
	\caption{Overall Algorithm}
	\label{alg1}
	
\end{algorithm}

%
%

\section{Anomaly Detection}
The entire crowd is often characterized by some dominant patterns, based on which, the entire set of trajectories is clustered. The anomalous trajectories, present throughout the crowded scene may belong to any one of these clusters but as a general property, will not have a substantial degree of belongingness to any of the clusters. The entire mechanism depends on two major tasks, as follows: Detecting Anomalies for each independent feature space followed by the selection of those trajectories that exhibit anomalous behaviour in most of the cases, using a voting mechanism. \\
Shannon Entropy has found widespread applications in numerous domains, with anomaly detection being one. The greatest advantage of this technique is that it allows the summarization of the feature distributions in the form of a single number. Our approach is based on the simple idea that an anomalous trajectory would exhibit higher levels of entropy when compared to normal trajectories. Instead of comparing the distances between the means of the cluster centres and trajectories as in previous work\cite{anjum2008multifeature}, we build a probability distribution using the distances between a trajectory and all of the cluster centres. The entropy of this probability distribution is evaluated and if it exceeds a threshold, it is classified as an anomaly. The threshold should be data adaptive and must adapt itself with the changing properties of the data.
\begin{center}
	$\mathnormal{C_i = [c_{1,i},c_{2,i},\ldots,c_{n,i}]}$\\[0.2cm]
	$\mathnormal{distvec_j = [distance(c_{1,i},f_j),\ldots,distance(c_{n,i},f_j)]}$\\[0.2cm]
\end{center}
$C_i$ represents the Cluster centres for a specific feature $i$ and the $distvec$ vector contains the distance measures between each of the cluster centre and trajectory $f_j$ . We further build the probability distribution $P_j = [p_{j,1}, p_{j,2}, \dots, p_{j,n}]$ where 
\begin{eqnarray}
p_{j,k} = \dfrac{distance(c_{k,i},f_j)}{\sum_{m=1}^{n}distance(c_{m,i}, f_j)}
\end{eqnarray}
An entropy measure is computed for each trajectory:
\begin{equation}
H_i = -\sum_{k=1}^{n} p_{i,k} \log_{a}p_{i,k}
\end{equation}
\begin{equation}
H = [H_1, H_2,\ldots, H_{numTraj}]
\end{equation}
Trajectories with an entropy value exceeding that of a threshold are marked anomalous for that feature. In a crowded scene, the changes in it's attributes occur randomly and most definitely. A particular section of the crowd can exhibit spatio-temporal changes in density and may also suddenly slow down or fasten up, thereby affecting individual feature parameters of the trajectories. Moreover, new trajectories that are introduced after a fixed interval of time may have similar features as a particular cluster but may exhibit one or more abnormalities due to it's late introduction. Therefore, we cannot club all trajectories marked as anomalous from the above stated procedure as our desired set of abnormalities. A simple voting mechanism sieves out those trajectories that are marked anomalous for majority of the cases.

\section{Results}
The method is tested on videos from two datasets, namely the Crowded scenes dataset used by Cheriyadat et al.\cite{cheriyadat2008detecting} to detect dominant motions in crowds and the UCF crowd dataset, first used by Ali et al.\cite{ali2007lagrangian}. To measure the efficiency of the method, we first identify all possible anomalous trajectories from the video and then, compare it with the classification test results. Since, the method involves the use of videos directly, we had to mark the anomalous trajectories in the actual video for the evaluation procedure. The results for three standard crowded videos from the mentioned datasets are tabulated as follows:

\begin{table}[htb]
	\begin{center}
		\begin{tabular}[h]{|c|c|c|c|c|}\hline
			Video&Precision&Recall&$f$-Score&Accuracy\\\hline
			Crowded Subway Exit&0.8258&0.9944&0.9023&96.31\%\\\hline
			Pilgrim Sequence&0.8221&0.9965&0.9009&98.15\%\\\hline
			Intersection Sequence&0.7287&0.9971&0.842&98.68\%\\\hline
		\end{tabular}
	\end{center}
	\label{tab1}
	\caption{Results on several Crowded Scene Videos}
\end{table}

The results indicate that this method exhibits excellent Specificity, i.e. the probability of classifying a normal trajectory as anomalous is extremely low. However, improvement can be achieved in the Sensitivity of the approach by improving the True Positive rate. It is to be noted that the method indicates almost all anomalous trajectories in the expected regions of interest with commendable accuracy. The graphical plots reveal the effective nature of the results produced. The plots as depicted in Figure 3 are from the Crowded subway exit sequence. The trajectories have been detected from the entire video sequence and thereafter, clustering has been done on the several feature-spaces as shown in Figures 3(a),3(c),3(e) and 3(g). The anomalous trajectories detected in each such feature space has been plotted in Figures 3(b),3(d),3(f) and 3(h). Following the voting mechanism, the final anomalous trajectories have been displayed as red curves with their origin points shown as blue dots in Figure 3(i). Figure 3(j) shows the overall crowded scene as being composed of the anomalous trajectories shown in red and the normal trajectories shown in blue. If the video is analyzed properly, one can find that the trajectories responsible for slowing down the crowd exiting the subway are closely represented by the ones detected as anomalous by the algorithm. These are in essence, the peripheral trajectories present together with the principal crowd flow that has been represented closely by the blue section in Figure 3(j).\\
\begin{table}[htb]
	\begin{center}
		\begin{tabular}[h]{|c|c|}\hline
			Method&Accuracy\\\hline
			Guo et. al.&96\%\\\hline
			Xu et. al.&87\%\\\hline
			Biswas et. al.&96.7\%\\\hline
			Proposed&98.68\%\\\hline
		\end{tabular}
	\end{center}
	\label{tab2}
	\caption{Comparison of Results}
\end{table}

The method performs well when compared with different state of the art methods. The overall accuracy has been used as the metric for comparison here. The Information Bottleneck based approach\cite{guo2014anomaly} only extracts a speed based feature to improve the shape analysis of trajectory data. This method shows an accuracy of about 96\% on their task-specific datasets. The other unsupervised methods, like the one based on hierarchical pattern discovery methods\cite{xu2014video}, although using a completely different approach; exhibit an accuracy of around 87\%. Other abnormality methods that use the property of sparsity in abnormal events\cite{biswas2017abnormality} exhibit an accuracy in the range of 88.71\% to 96.7\%. 
\section{Conclusion}
This paper stresses on the need for understanding crowd dynamics better and presents an unsupervised mechanism to detect anomalous trajectories. The method is an application-ready one that itself generates trajectories from a video using a multi-object tracker and then cluster them based on multiple independent features. The use of multiple features for determining the clusters and the anomalies is based on the fact that an anomalous trajectory may posses similarity with a dominant pattern in one aspect, but differs significantly in a majority of aspects. A trajectory that may be similar to most trajectories in terms of mean location and position may cause disturbance in the scene due to its unnatural speed. This has been taken care of by using multiple features to detect the overall anomalies. The use of Shannon Entropy provides a novel approach to determine the anomalies, considering the fact that a probability distribution is developed using the distances from all cluster centres and not only the specific cluster with which the trajectory is associated. An anomalous trajectory is unlikely to belong to any specific cluster to a significant degree, thereby maximizing entropy in the probability distribution. The proposed approach yields excellent results on the chosen crowd videos. This work can be made efficient by developing a substantially large dataset that demarcates abnormal trajectories where the trajectories are represented as a time series as used here. Trajectory representation such as this has been used to evaluate crowd flow segmentation but here it has been put to use for abnormality detection. This lends this approach the added advantage of detecting the specific areas in the scene that contribute majorly to disturbance. Finally, this work may find extensive use in improving surveillance methods, better public space design, efficient event organization and possibly, even in tracking rogue naval and air routes. This work can be improved by making it real-time and also by generalizing the Entropy measure that could classify the anomalies optimally. 

\begin{figure}[htbp]
	
	\subfigure[Clustering based on the density feature]
	{\label{fig:edge-a}
		\includegraphics[scale=0.35]{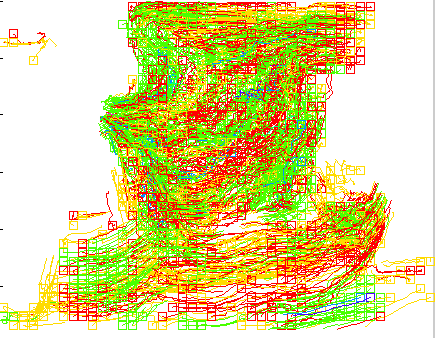}}
	\subfigure[Anomalous trajectories based on Density feature]
	{\label{fig:edge-b}
		\includegraphics[scale=0.35]{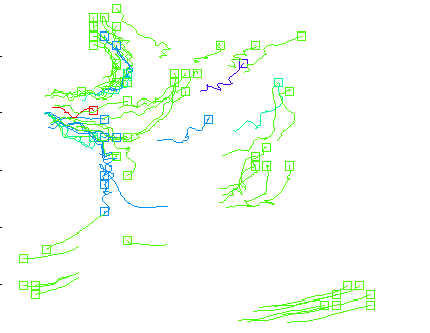}}\\
	
	\subfigure[Clustering based on the Shape feature]
	{\label{fig:edge-c}
		\includegraphics[scale=0.35]{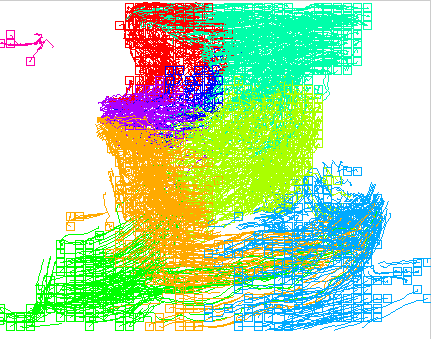}}
	\subfigure[Anomalous trajectories based on Shape feature]
	{\label{fig:edge-d}
		\includegraphics[scale=0.35]{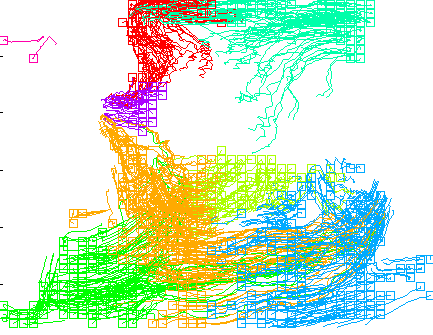}}\\
	
	\subfigure[Clustering based on the Mean Position]
	{\label{fig:edge-e}
		\includegraphics[scale=0.35]{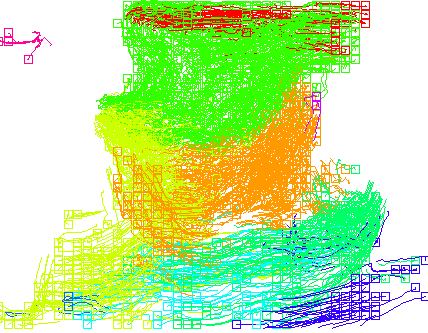}}
	\subfigure[Anomalous trajectories based on Mean Position]
	{\label{fig:edge-f}
		\includegraphics[scale=0.35]{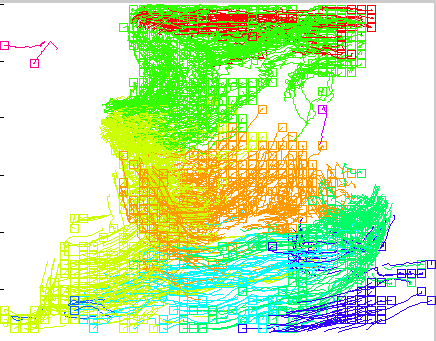}}\\
	
	\subfigure[Clustering based on the Standard Deviation]
	{\label{fig:edge-g}
		\includegraphics[scale=0.35]{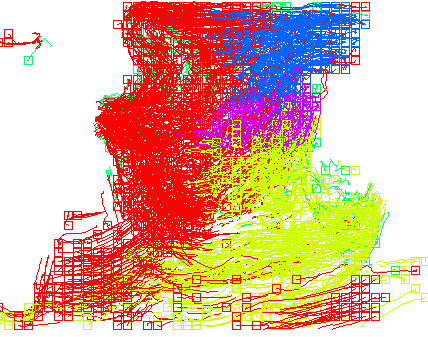}}
	\subfigure[Anomalous trajectories based on Standard Deviation]
	{\label{fig:edge-h}
		\includegraphics[scale=0.35]{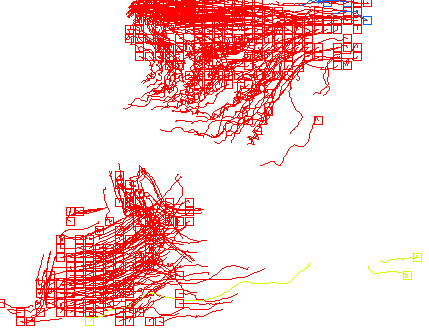}}\\
	
	\subfigure[Overall Anomalies]
	{\label{fig:edge-i}
		\includegraphics[scale=0.35]{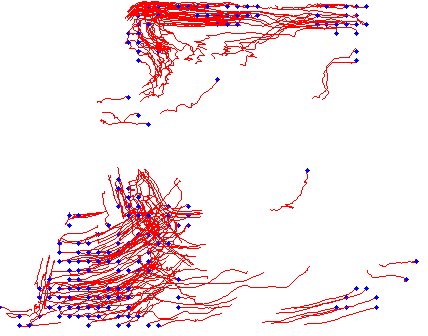}}
	\subfigure[Overall Scene]
	{\label{fig:edge-j}
		\includegraphics[scale=0.35]{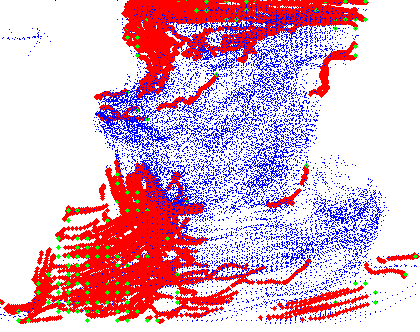}}
	
	\caption{Different Clusterings and Anomalous Trajectory Classification}
	\label{fig:edge}
\end{figure}

\bibliographystyle{IEEEtran}

\end{document}